\documentclass{article} 
\usepackage{iclr2023_conference,times}
\usepackage{graphicx}

\usepackage{amsmath,amsfonts,bm}

\def\eqref#1{equation~\ref{#1}}

\def\1{\bm{1}}

\DeclareMathAlphabet{\mathsfit}{\encodingdefault}{\sfdefault}{m}{sl}
\SetMathAlphabet{\mathsfit}{bold}{\encodingdefault}{\sfdefault}{bx}{n}

\usepackage{hyperref}
\usepackage{url}
\usepackage[font=small,skip=0pt]{caption}

\title{Synthetic Data Generator  for  \protect\\Adaptive Interventions in Global Health}

\author{\textbf{Aditya Rastogi, Juan Francisco Garamendi, Ana Fernández del Río, Anna Guitart,} \\
\textbf{Moiz Hassan Khan, Dexian Tang \& África Periáñez} \\ 
benshi.ai\\
Barcelona, Spain \\
\texttt{\{aditya,juanfran,ana,guitart,moiz,dexian,africa\}@benshi.ai} \\
}

\iclrfinalcopy 

\begin{document}

\maketitle
\linepenalty=100

\begin{abstract}


Artificial Intelligence and digital health have the potential to transform global health. However, having access to representative data to test and validate algorithms in realistic production environments is essential. We introduce \emph{HealthSyn}, an open-source synthetic data generator of user behavior for testing reinforcement learning algorithms in the context of mobile health interventions. The generator utilizes Markov processes to generate diverse user actions, with individual user behavioral patterns that can change in reaction to personalized interventions (i.e., reminders, recommendations, and incentives). These actions are translated into actual logs using an ML-purposed data schema specific to the mobile health application functionality included with \emph{HealthKit}, and open-source SDK. The logs can be fed to pipelines to obtain user metrics. The generated data, which is based on real-world behaviors and simulation techniques, can be used to develop, test, and evaluate, both ML algorithms in research and end-to-end operational RL-based intervention delivery frameworks.

\end{abstract}



\section{Introduction}
\label{sec:intro}
Digital tools have become essential for frontline health workers (FHWs) everywhere, particularly in resource-poor settings where they are often designed to mitigate the effects of the lack of resources or infrastructure~\citep{marsch2021digital,overdijkink2018usability}. From capacity building and communication to patient management and triage assistance or medical supply delivery, there is an array of mobile health solutions to help workers provide the best care possible to their communities~\citep{overdijkink2018usability,hosny2019artificial,wahl2018artificial,forero2018application}. 

These tools provide an excellent opportunity for personalization through adaptive technologies, using the rich information about provider and patient behavior and outcomes contained in the records generated through the use of digital applications~\citep{olaniyi2022user,benshi2021forecasting,benshi2021recommendation,Katsaris2021}. This allows us to provide additional support by adapting the content to each FHW's and patient's evolving needs and by delivering timely reminders, suggestions, or personalized incentives specifically designed to work for them at precisely that time. We thus increase the chances of success by adopting an adaptive, contextualized approach to innovation that allows for rapid deployment cycles. Big tech has excelled at this iterative model, although optimizing for engagement and monetization rather than well-being.

Statistical analysis and machine learning (ML) models play a crucial role in using the information available to understand and predict individual FHW or patient behavior in order to design the best interventions for each use case~\citep{hosny2019artificial,wahl2018artificial}. Reinforcement learning (RL), a subset of these methodologies, additionally provides the algorithmic framework for making decisions about when and which interventions to send~\citep{Yom17, Forman19, Liao20, Wang2021, Trella22}. It allows us to fine-tune the system depending on how much weight on knowledge extraction (i.e., gathering data that will allow statistical inference with enough power) or on optimization (i.e., using all information to make the best choice for each individual even if it hinders statistical significance) we want to put. 

Synthetic data is becoming increasingly important due to concerns about reproducibility and establishing baselines~\citep{Jordon2020}. Healthcare data is extraordinarily sensitive, and privacy through appropriate de-identification is a crucial step (largely yet to be overcome) before any data can be shared for analysis and ML ~\citep{Yoon2020, Jordon2021}. Adaptive technologies involve complex feedback cycles and long-term effects,  so having a way to simulate user behavior, both in the absence of interventions and in their reactions to them, becomes essential for safe development and testing before the deployment with real subjects.

We introduce \emph{HealthSyn}~\citep{healthsyn}, an open-source library for mobile health data generation, and describe its components and methodology. It can be used for ML model development and can be coupled to algorithmic decision-making on interventions to which the user will react, thus providing an end-to-end simulation environment for RL-based interventions in mobile health. User behavioral logs from mobile health applications are tracked using \emph{HealthKit}~\citep{sdk}, an open-source Software Developer Kit (SDK) for mobile health. It is responsible for the tracking, labeling, and organization of the logs in a data schema optimized for ML and intervention purposes. \emph{HealthKit} tracks general in-app users' behavior (actions, sessions, etc.) and additional events organized based on the type of content (patient management, capacity building, etc.). It also delivers the interventions and tracks the user's reaction to them (e.g. whether notifications were opened, discarded, or blocked). It distinguishes between online and offline activity, which is particularly important when working with global health applications, as they are often used without an internet connection, impacting both user behavior and the interventions that will work best for them.

\subsection{Our contribution}
\label{sec:our-contribution}

To the best of our knowledge, \emph{HealthSyn} is the first synthetic log generation library openly available. Complemented by the open \emph{HealthKit} SDK, responsible for most of the heavy lifting regarding data tracking, labeling, and organization, it provides a simulation environment for RL-based interventions for mobile health. Although other behavioral simulators exist (see appendix \ref{app:related-work} \footnote{Related works have not been included in the main paper for the sake of space, but the interested reader can find this section as an appendix.}), they focus on modeling subject daily activity (e.g., sleep hours, or alcohol consumption). The simulation environment presented here generates the information as if generated by app use. It is well-suited explicitly for digital tools designed to support FHW in low- and middle-income countries. 

\section{Mobile Health Log Synthetic Generator Architecture}
\label{sec:synthetic-data}


\begin{figure}[ht]
\begin{center}
\includegraphics[width=\linewidth]{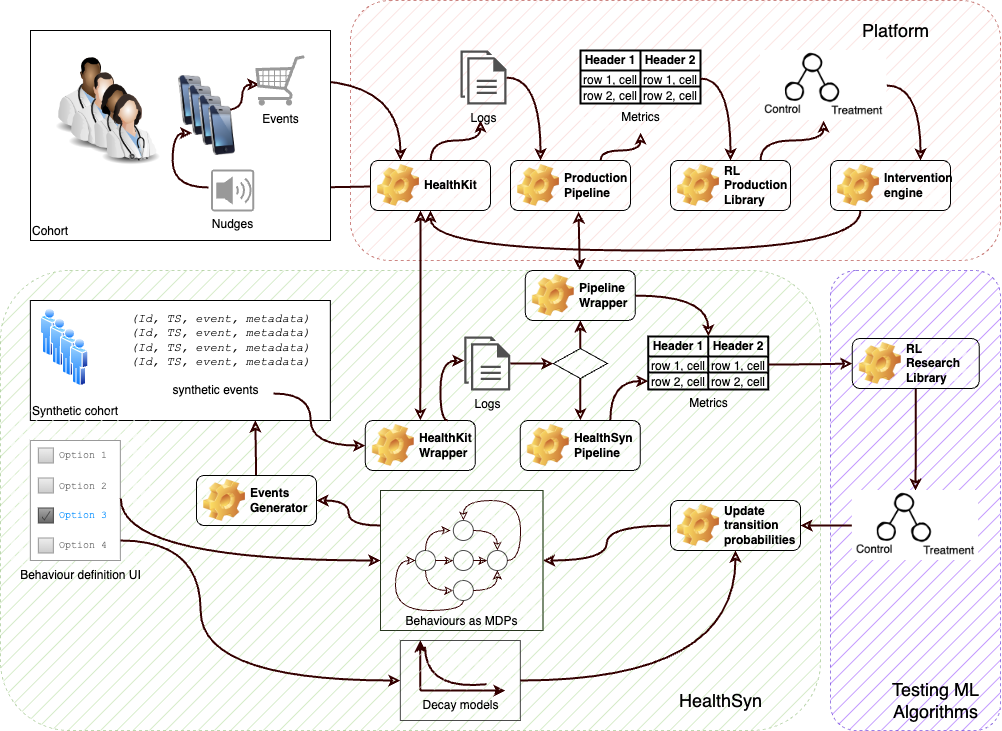}
\end{center}
\caption{Overview of the production (modules involved in the delivery of interventions to real users in production settings, shaded in orange) and simulation (modules that mimic synthetic users and their reactions to interventions, shaded in green) environments. Modules included with \emph{HealthSyn} are highlighted, and the \emph{HealthKit} is included as the interface between users and the platform.}
\label{fig:overall_process}
\end{figure}

\emph{HealthSyn} is made up of three components: the \emph{environment definition} and \emph{event generator} modules, and a collection of \emph{external wrappers} to link those modules to \emph{HealthKit}, data pipelines, and RL-based nudge delivery system. These components allow for the simulation of an end-to-end intervention delivery framework as depicted in Figure \ref{fig:overall_process}, which shows the production system (described in more detail in appendix \ref{app:ml-adaptive-intervention}) shaded in orange, and how this is emulated using \emph{HealthSyn} in green.

\begin{figure}[ht]
    \begin{center}
        \includegraphics[width=0.8\linewidth]{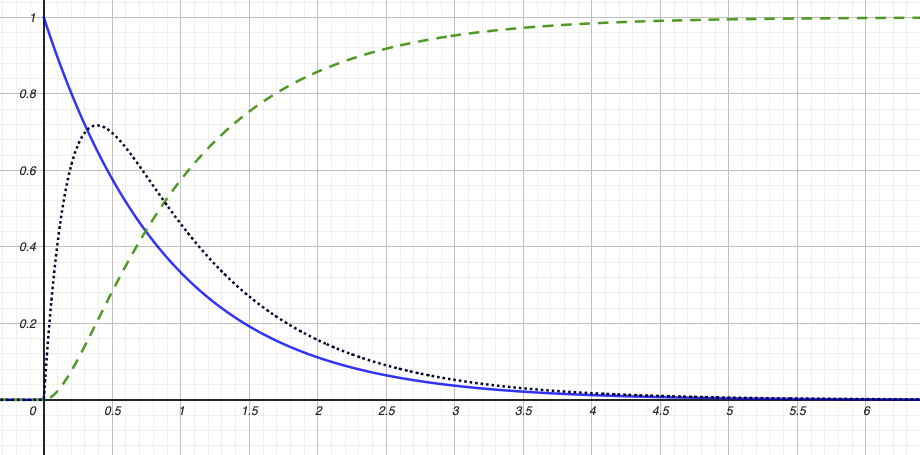}
    \end{center}
    \caption{Decay functions used to model user reaction to nudges. The x-axis represents the number of nudges received.
    The solid blue line represents a response inversely proportional to the interventions (nudging has a negative effect). The black dotted line is characteristic of a positive reaction to the intervention until fatigue wears off the effect. The dashed green line represents a positive reaction that does not decay, but has decreasing returns that saturate the effect at a certain level.}
    \label{fig:modelBehaviour}
\end{figure}

\subsection{Environment Definition}

The \emph{environment definition} module is composed of submodules responsible for determining user behavior (both unaltered and in response to nudging) and the context (i.e., the set of traits that will be considered as defining their state) and rewards for the contextual or restless bandits from the RL-component with which \emph{HealthSyn} can be coupled for adaptive intervention simulation. 

User behavior is modeled using a Markov processes. A key component of this model is the transition probability matrix, which can be defined differently for each user or context, or be the same for the entire population. This matrix serves as a baseline for the users' behavior, and it establishes the behavior in the absence of interventions. 

User response to nudges is defined using a decay model that takes into account the number of nudges received on the last $d$ days (where $d$ is another parameter to be set for the environment). The decay model is a linear combination of three functions, each representing different types of user behavior. These are represented in Figure \ref{fig:modelBehaviour}, with parameters that can be determined through a fitting process using real-world historical data to ensure that the model accurately reflects users' actual behavior. One function (in solid blue) models an effect that is maximum when no interventions are received (the nudging negatively impacts users whose behavior is modeled by it exclusively). Another function (in dotted black) represents an initial positive effect that eventually whines to zero. The final function (in dashed green) models an effect that builds up with increasing returns per nudge until it saturates at a constant level. Note that the combination of these can model complex user reactions to nudging, including long-lasting, novelty, and fatigue effects. See appendix \ref{appendix:maths} for a formal definition of the functions.

\subsection{Synthetic In-App Activity Generation}

\begin{figure}[ht]
    \begin{center}
        \includegraphics[width=\linewidth]{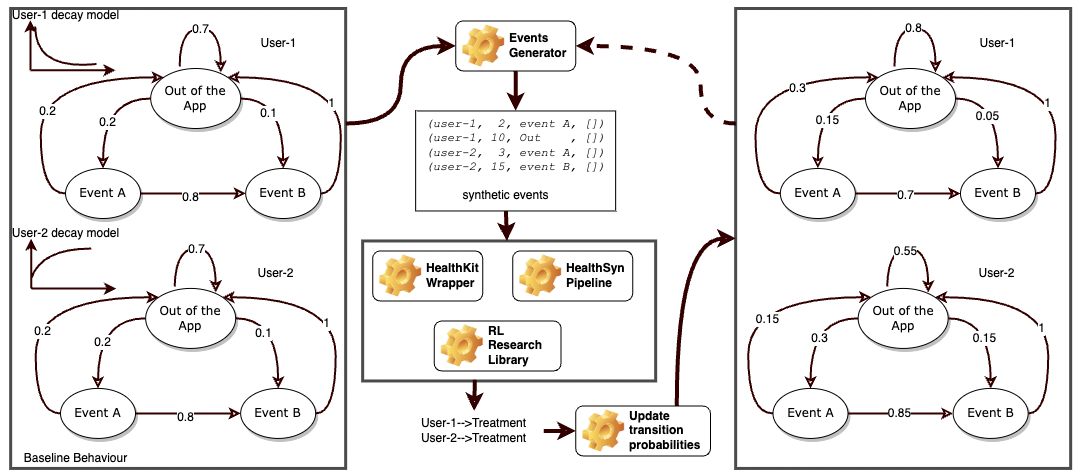}
    \end{center}
    \caption{Overview of a single iteration in the simulation. The baseline behavior as defined in the \emph{environment definition} and used by the \emph{event generator} module to create a comprehensive list of in-app events or actions. These events are then translated into metrics that are used by the RL algorithm for intervention decisions, which in turn impacts user transition probabilities according to the user decay model. Dashed arrows represent input for the next iteration.}
    \label{fig:simMarkov}
\end{figure}

The \emph{event generator} module generates user in-app actions  or events as a collection of tuples, with each tuple representing an (id, timestamp, action, and metadata). Users are modeled as Markov processes; the generated actions are the states reached through the process. 


Figure \ref{fig:simMarkov} illustrates the dynamic changes in the transition matrix for two users due to online interventions as the simulation progresses. In this illustration, we observe that both users start with the same baseline transition matrix and receive an intervention at the last decision point. However, the matrix evolves differently for each user based on each particular decay model. As shown in Figure \ref{fig:modelBehaviour}, 'user-1' has an adverse reaction to interventions and is modeled as the blue function, while 'user-2' has a positive response and is modeled as the dashed green line. The 'action generator' module adjusts the transition matrix accordingly, with probabilities of specific actions decreasing and probabilities of transitioning to the 'Out of the app' state increasing for 'user-1' and the opposite occurring for 'user-2'.
For users who do not receive an intervention, the transition probabilities will not change.

\subsection{External Wrapper}

 This module translates the simulated user actions into actual logs and metrics. It serves as the bridge between the simulated world and the real world, and it is composed of three sub-modules. The \emph{HealthKit Wrapper} reads and interprets the actions and then uses the SDK to generate the associated logs in the SDK schema. The \emph{Pipeline Wrapper} reads the logs generated by the SDK wrapper, uses an existing data pipeline to compute the metrics, and saves them into a database. Finally, the \emph{RL Wrapper} converts these metrics into context and rewards (according to the environment definition) that can be directly fed to the chosen RL algorithm. It also feeds back information on intervention decisions to the action generator module.





\section{Summary and Conclusions}
\label{sec:conclusions}

We have proposed a workflow that offers a systematic approach for evaluating RL algorithms for adaptive interventions in mobile health using synthetic data. It uses \emph{HealthSyn}, an open-source library to simulate the behavior of FHWs in digital tools, and \emph{HealthKit}, an open-source SDK which can track and label their logs in a data schema optimized for ML. 

\bibliography{iclr2023_synloggen}

\begin{thebibliography}{35}
\providecommand{\natexlab}[1]{#1}
\providecommand{\url}[1]{\texttt{#1}}
\expandafter\ifx\csname urlstyle\endcsname\relax
  \providecommand{\doi}[1]{doi: #1}\else
  \providecommand{\doi}{doi: \begingroup \urlstyle{rm}\Url}\fi

\bibitem[Alaa et~al.(2021{\natexlab{a}})Alaa, Chan, and van~der
  Schaar]{Alaa2021a}
Ahmed Alaa, Alex~James Chan, and Mihaela van~der Schaar.
\newblock Generative time-series modeling with fourier flows.
\newblock In \emph{International Conference on Learning Representations},
  2021{\natexlab{a}}.
\newblock URL \url{https://openreview.net/forum?id=PpshD0AXfA}.

\bibitem[Alaa \& van~der Schaar(2019)Alaa and van~der Schaar]{Alaa2019}
Ahmed~M. Alaa and Mihaela van~der Schaar.
\newblock Attentive state-space modeling of disease progression.
\newblock In H.~Wallach, H.~Larochelle, A.~Beygelzimer, F.~d\textquotesingle
  Alch\'{e}-Buc, E.~Fox, and R.~Garnett (eds.), \emph{Advances in Neural
  Information Processing Systems}, volume~32. Curran Associates, Inc., 2019.
\newblock URL
  \url{https://proceedings.neurips.cc/paper/2019/file/1d0932d7f57ce74d9d9931a2c6db8a06-Paper.pdf}.

\bibitem[Alaa et~al.(2021{\natexlab{b}})Alaa, van Breugel, Saveliev, and
  van~der Schaar]{Alaa2021}
Ahmed~M. Alaa, Boris van Breugel, Evgeny Saveliev, and Mihaela van~der Schaar.
\newblock How faithful is your synthetic data? sample-level metrics for
  evaluating and auditing generative models, February 2021{\natexlab{b}}.

\bibitem[benshi.ai(2023{\natexlab{a}})]{healthsyn}
benshi.ai.
\newblock Health{S}yn.
\newblock HealthSyn GitHub repository, 2023{\natexlab{a}}.
\newblock URL \url{https://github.com/benshi-ai/healthsyn}.

\bibitem[benshi.ai(2023{\natexlab{b}})]{sdk}
benshi.ai.
\newblock Health{K}it.
\newblock NPM package, 2023{\natexlab{b}}.
\newblock URL \url{https://www.npmjs.com/package/@benshi.ai/js-sdk}.

\bibitem[Brockman et~al.(2016)Brockman, Cheung, Pettersson, Schneider,
  Schulman, Tang, and Zaremba]{Brockman2016}
Greg Brockman, Vicki Cheung, Ludwig Pettersson, Jonas Schneider, John Schulman,
  Jie Tang, and Wojciech Zaremba.
\newblock Openai gym, 2016.

\bibitem[Forero et~al.(2018)Forero, Nahidi, De~Costa, Mohsin, Fitzgerald,
  Gibson, McCarthy, and Aboagye-Sarfo]{forero2018application}
Roberto Forero, Shizar Nahidi, Josephine De~Costa, Mohammed Mohsin, Gerry
  Fitzgerald, Nick Gibson, Sally McCarthy, and Patrick Aboagye-Sarfo.
\newblock Application of four-dimension criteria to assess rigour of
  qualitative research in emergency medicine.
\newblock \emph{BMC health services research}, 18\penalty0 (1):\penalty0 1--11,
  2018.

\bibitem[Forman et~al.(2019)Forman, Kerrigan, Butryn, Juarascio, Manasse,
  Ontañón, Dallal, Crochiere, and Moskow]{Forman19}
Evan~M. Forman, Stephanie~G. Kerrigan, Meghan~L. Butryn, Adrienne~S. Juarascio,
  Stephanie~M. Manasse, Santiago Ontañón, Diane~H. Dallal, Rebecca~J.
  Crochiere, and Danielle Moskow.
\newblock Can the artificial intelligence technique of reinforcement learning
  use continuously-monitored digital data to optimize treatment for weight
  loss?
\newblock \emph{Journal of behavioral medicine}, 42:\penalty0 276, 4 2019.
\newblock ISSN 15733521.
\newblock \doi{10.1007/S10865-018-9964-1}.
\newblock URL \url{/pmc/articles/PMC6524648/
  /pmc/articles/PMC6524648/?report=abstract
  https://www.ncbi.nlm.nih.gov/pmc/articles/PMC6524648/}.

\bibitem[Guitart et~al.(2021{\natexlab{a}})Guitart, del Río, África
  Periáñez, and Bellhouse]{benshi2021forecasting}
Anna Guitart, Ana~Fernández del Río, África Periáñez, and Lauren
  Bellhouse.
\newblock Midwifery {L}earning and {F}orecasting: {P}redicting {C}ontent
  {D}emand with {U}ser-{G}enerated {L}ogs, 2021{\natexlab{a}}.

\bibitem[Guitart et~al.(2021{\natexlab{b}})Guitart, Heydari, Olaleye, Ljubicic,
  del R{\'\i}o, Peri{\'a}{\~n}ez, and Bellhouse]{benshi2021recommendation}
Anna Guitart, Afsaneh Heydari, Eniola Olaleye, Jelena Ljubicic,
  Ana~Fern{\'a}ndez del R{\'\i}o, {\'A}frica Peri{\'a}{\~n}ez, and Lauren
  Bellhouse.
\newblock A recommendation system to enhance midwives' capacities in low-income
  countries.
\newblock \emph{arXiv preprint arXiv:2111.01786}, 2021{\natexlab{b}}.

\bibitem[Hosny \& Aerts(2019)Hosny and Aerts]{hosny2019artificial}
Ahmed Hosny and Hugo~JWL Aerts.
\newblock Artificial intelligence for global health.
\newblock \emph{Science}, 366\penalty0 (6468):\penalty0 955--956, 2019.

\bibitem[Ie et~al.(2019)Ie, wei Hsu, Mladenov, Jain, Narvekar, Wang, Wu, and
  Boutilier]{Ie2019b}
Eugene Ie, Chih wei Hsu, Martin Mladenov, Vihan Jain, Sanmit Narvekar, Jing
  Wang, Rui Wu, and Craig Boutilier.
\newblock Recsim: A configurable simulation platform for recommender systems,
  2019.

\bibitem[Jarrett et~al.(2021)Jarrett, Bica, and van~der Schaar]{Jarrett2021}
Daniel Jarrett, Ioana Bica, and Mihaela van~der Schaar.
\newblock Time-series generation by contrastive imitation.
\newblock In M.~Ranzato, A.~Beygelzimer, Y.~Dauphin, P.S. Liang, and J.~Wortman
  Vaughan (eds.), \emph{Advances in Neural Information Processing Systems},
  volume~34, pp.\  28968--28982. Curran Associates, Inc., 2021.
\newblock URL
  \url{https://proceedings.neurips.cc/paper/2021/file/f2b4053221961416d47d497814a8064f-Paper.pdf}.

\bibitem[Jordon et~al.(2018{\natexlab{a}})Jordon, Yoon, and van~der
  Schaar]{Jordon2018}
James Jordon, Jinsung Yoon, and Mihaela van~der Schaar.
\newblock Measuring the quality of synthetic data for use in competitions.
\newblock \emph{CoRR}, abs/1806.11345, 2018{\natexlab{a}}.
\newblock URL \url{http://arxiv.org/abs/1806.11345}.

\bibitem[Jordon et~al.(2018{\natexlab{b}})Jordon, Yoon, and van~der
  Schaar]{Jordon2018a}
James Jordon, Jinsung Yoon, and Mihaela van~der Schaar.
\newblock Pate-gan: Generating synthetic data with differential privacy
  guarantees.
\newblock In \emph{International Conference on Learning Representations},
  2018{\natexlab{b}}.

\bibitem[Jordon et~al.(2019)Jordon, Yoon, and van~der Schaar]{Jordon2019}
James Jordon, Jinsung Yoon, and Mihaela van~der Schaar.
\newblock Differentially private bagging: Improved utility and cheaper privacy
  than subsample-and-aggregate.
\newblock In H.~Wallach, H.~Larochelle, A.~Beygelzimer, F.~d\textquotesingle
  Alch\'{e}-Buc, E.~Fox, and R.~Garnett (eds.), \emph{Advances in Neural
  Information Processing Systems}, volume~32. Curran Associates, Inc., 2019.
\newblock URL
  \url{https://proceedings.neurips.cc/paper/2019/file/5dec707028b05bcbd3a1db5640f842c5-Paper.pdf}.

\bibitem[Jordon et~al.(2020)Jordon, Wilson, and van~der Schaar]{Jordon2020}
James Jordon, Alan Wilson, and Mihaela van~der Schaar.
\newblock Synthetic data: Opening the data floodgates to enable faster, more
  directed development of machine learning methods.
\newblock \emph{CoRR}, abs/2012.04580, 2020.
\newblock URL \url{https://arxiv.org/abs/2012.04580}.

\bibitem[Jordon et~al.(2021)Jordon, Jarrett, Saveliev, Yoon, Elbers, Thoral,
  Ercole, Zhang, Belgrave, and van~der Schaar]{Jordon2021}
James Jordon, Daniel Jarrett, Evgeny Saveliev, Jinsung Yoon, Paul Elbers,
  Patrick Thoral, Ari Ercole, Cheng Zhang, Danielle Belgrave, and Mihaela
  van~der Schaar.
\newblock Hide-and-seek privacy challenge: Synthetic data generation vs.
  patient re-identification.
\newblock In Hugo~Jair Escalante and Katja Hofmann (eds.), \emph{Proceedings of
  the NeurIPS 2020 Competition and Demonstration Track}, volume 133 of
  \emph{Proceedings of Machine Learning Research}, pp.\  206--215. PMLR, 06--12
  Dec 2021.
\newblock URL \url{https://proceedings.mlr.press/v133/jordon21a.html}.

\bibitem[Katsaris et~al.(2021)Katsaris, , and and]{Katsaris2021}
Iraklis Katsaris, , and Nikolas~Vidakis and.
\newblock Adaptive e-learning systems through learning styles: A review of the
  literature.
\newblock \emph{Advances in Mobile Learning Educational Research}, 1\penalty0
  (2):\penalty0 124--145, 2021.
\newblock \doi{10.25082/amler.2021.02.007}.

\bibitem[Liao et~al.(2020)Liao, Greenewald, Klasnja, and Murphy]{Liao20}
Peng Liao, Kristjan Greenewald, Predrag Klasnja, and Susan Murphy.
\newblock Personalized heartsteps: A reinforcement learning algorithm for
  optimizing physical activity.
\newblock \emph{Proceedings of the ACM on interactive, mobile, wearable and
  ubiquitous technologies}, 4, 3 2020.
\newblock ISSN 24749567.
\newblock \doi{10.1145/3381007}.
\newblock URL \url{/pmc/articles/PMC8439432/
  /pmc/articles/PMC8439432/?report=abstract
  https://www.ncbi.nlm.nih.gov/pmc/articles/PMC8439432/}.

\bibitem[Marsch(2021)]{marsch2021digital}
Lisa~A Marsch.
\newblock Digital health data-driven approaches to understand human behavior.
\newblock \emph{Neuropsychopharmacology}, 46\penalty0 (1):\penalty0 191--196,
  2021.

\bibitem[Nguyen et~al.(2019)Nguyen, Karim, Vu, Schl{\"o}tterer, and
  Granitzer]{Nguyen2019}
Van~Bach Nguyen, Belaid~Mohamed Karim, Bao~Long Vu, J{\"o}rg Schl{\"o}tterer,
  and Michael Granitzer.
\newblock Policy learning for malaria control.
\newblock \emph{ArXiv}, abs/1910.08926, 2019.

\bibitem[Olaniyi et~al.(2022)Olaniyi, del R{\'\i}o, Peri{\'a}{\~n}ez, and
  Bellhouse]{olaniyi2022user}
Babaniyi~Yusuf Olaniyi, Ana~Fern{\'a}ndez del R{\'\i}o, {\'A}frica
  Peri{\'a}{\~n}ez, and Lauren Bellhouse.
\newblock User engagement and churn in mobile health applications.
\newblock \emph{In Proceedings of 2022 ACM KDD Workshop on Applied Data Science
  for Healthcare (DSHealth 2022)}, 2022.

\bibitem[Overdijkink et~al.(2018)Overdijkink, Velu, Rosman, Van~Beukering, Kok,
  and Steegers-Theunissen]{overdijkink2018usability}
Sanne~B Overdijkink, Adeline~V Velu, Ageeth~N Rosman, Monique~DM Van~Beukering,
  Marjolein Kok, and Regine~PM Steegers-Theunissen.
\newblock The usability and effectiveness of mobile health technology--based
  lifestyle and medical intervention apps supporting health care during
  pregnancy: systematic review.
\newblock \emph{JMIR mHealth and uHealth}, 6\penalty0 (4):\penalty0 e8834,
  2018.

\bibitem[Remy \& Bent(2020)Remy and Bent]{Remy2020}
Sekou~L. Remy and Oliver Bent.
\newblock A global health gym environment for rl applications.
\newblock In Hugo~Jair Escalante and Raia Hadsell (eds.), \emph{Proceedings of
  the NeurIPS 2019 Competition and Demonstration Track}, volume 123 of
  \emph{Proceedings of Machine Learning Research}, pp.\  253--261. PMLR, 08--14
  Dec 2020.
\newblock URL \url{https://proceedings.mlr.press/v123/remy20a.html}.

\bibitem[Santana et~al.(2020)Santana, Melo, Camargo, Brandao, Soares, Oliveira,
  and Caetano]{Mars20}
Marlesson~R.O. Santana, Luckeciano~C. Melo, Fernando~H.F. Camargo, Bruno
  Brandao, Anderson Soares, Renan~M. Oliveira, and Sandor Caetano.
\newblock Mars-gym: A gym framework to model, train, and evaluate recommender
  systems for marketplaces.
\newblock \emph{2020 International Conference on Data Mining Workshops
  (ICDMW)}, 2020-November:\penalty0 189--197, 11 2020.
\newblock ISSN 23759259.
\newblock \doi{10.1109/ICDMW51313.2020.00035}.

\bibitem[Singh et~al.(2020)Singh, Halpern, Thain, Christakopoulou, Chi, Chen,
  Beutel, and Research]{Singh2020}
Ashudeep Singh, Yoni Halpern, Nithum Thain, Konstantina Christakopoulou, Ed~H
  Chi, Jilin Chen, Alex Beutel, and Google Research.
\newblock Building healthy recommendation sequences for everyone: A safe
  reinforcement learning approach.
\newblock In \emph{ACM RecSys 2020}, 9 2020.

\bibitem[Trella et~al.(2022)Trella, Zhang, Nahum-Shani, Shetty, Doshi-Velez,
  and Murphy]{Trella22}
Anna~L. Trella, Kelly~W. Zhang, Inbal Nahum-Shani, Vivek Shetty, Finale
  Doshi-Velez, and Susan~A. Murphy.
\newblock Designing reinforcement learning algorithms for digital
  interventions: Pre-implementation guidelines.
\newblock \emph{Algorithms 2022, Vol. 15, Page 255}, 15:\penalty0 255, 7 2022.
\newblock ISSN 1999-4893.
\newblock \doi{10.3390/A15080255}.
\newblock URL \url{https://www.mdpi.com/1999-4893/15/8/255/htm
  https://www.mdpi.com/1999-4893/15/8/255}.

\bibitem[van Breugel et~al.(2021)van Breugel, Kyono, Berrevoets, and van~der
  Schaar]{Breugel2021}
Boris van Breugel, Trent Kyono, Jeroen Berrevoets, and Mihaela van~der Schaar.
\newblock Decaf: Generating fair synthetic data using causally-aware generative
  networks.
\newblock In M.~Ranzato, A.~Beygelzimer, Y.~Dauphin, P.S. Liang, and J.~Wortman
  Vaughan (eds.), \emph{Advances in Neural Information Processing Systems},
  volume~34, pp.\  22221--22233. Curran Associates, Inc., 2021.
\newblock URL
  \url{https://proceedings.neurips.cc/paper/2021/file/ba9fab001f67381e56e410575874d967-Paper.pdf}.

\bibitem[Wahl et~al.(2018)Wahl, Cossy-Gantner, Germann, and
  Schwalbe]{wahl2018artificial}
Brian Wahl, Aline Cossy-Gantner, Stefan Germann, and Nina~R Schwalbe.
\newblock Artificial intelligence ({AI}) and global health: how can {AI}
  contribute to health in resource-poor settings?
\newblock \emph{BMJ global health}, 3\penalty0 (4):\penalty0 e000798, 2018.

\bibitem[Wang et~al.(2021)Wang, Zhang, Kröse, and van Hoof]{Wang2021}
Shihan Wang, Chao Zhang, Ben Kröse, and Herke van Hoof.
\newblock Optimizing adaptive notifications in mobile health interventions
  systems: Reinforcement learning from a data-driven behavioral simulator.
\newblock \emph{Journal of Medical Systems}, 45:\penalty0 1--8, 12 2021.
\newblock ISSN 1573689X.
\newblock \doi{10.1007/S10916-021-01773-0/FIGURES/4}.
\newblock URL
  \url{https://link.springer.com/article/10.1007/s10916-021-01773-0}.

\bibitem[Yom-Tov et~al.(2017)Yom-Tov, Feraru, Kozdoba, Mannor, Tennenholtz, and
  Hochberg]{Yom17}
Elad Yom-Tov, Guy Feraru, Mark Kozdoba, Shie Mannor, Moshe Tennenholtz, and
  Irit Hochberg.
\newblock Encouraging physical activity in patients with diabetes: Intervention
  using a reinforcement learning system.
\newblock \emph{J Med Internet Res 2017;19(10):e338
  https://www.jmir.org/2017/10/e338}, 19:\penalty0 e7994, 10 2017.
\newblock ISSN 14388871.
\newblock \doi{10.2196/JMIR.7994}.
\newblock URL \url{https://www.jmir.org/2017/10/e338}.

\bibitem[Yoon et~al.(2019)Yoon, Jarrett, and van~der Schaar]{Yoon2019}
Jinsung Yoon, Daniel Jarrett, and Mihaela van~der Schaar.
\newblock Time-series generative adversarial networks.
\newblock In H.~Wallach, H.~Larochelle, A.~Beygelzimer, F.~d\textquotesingle
  Alch\'{e}-Buc, E.~Fox, and R.~Garnett (eds.), \emph{Advances in Neural
  Information Processing Systems}, volume~32. Curran Associates, Inc., 2019.
\newblock URL
  \url{https://proceedings.neurips.cc/paper/2019/file/c9efe5f26cd17ba6216bbe2a7d26d490-Paper.pdf}.

\bibitem[Yoon et~al.(2020)Yoon, Drumright, and van~der Schaar]{Yoon2020}
Jinsung Yoon, Lydia~N. Drumright, and Mihaela van~der Schaar.
\newblock Anonymization through data synthesis using generative adversarial
  networks (ads-gan).
\newblock \emph{IEEE Journal of Biomedical and Health Informatics}, 24\penalty0
  (8):\penalty0 2378--2388, 2020.
\newblock \doi{10.1109/JBHI.2020.2980262}.

\bibitem[Zou(2021)]{Zou2021}
Lixin Zou.
\newblock Data-efficient reinforcement learning for malaria control.
\newblock In Zhi-Hua Zhou (ed.), \emph{Proceedings of the Thirtieth
  International Joint Conference on Artificial Intelligence, {IJCAI-21}}, pp.\
  507--513. International Joint Conferences on Artificial Intelligence
  Organization, 8 2021.
\newblock \doi{10.24963/ijcai.2021/71}.
\newblock URL \url{https://doi.org/10.24963/ijcai.2021/71}.
\newblock Main Track.

\end{thebibliography}
\bibliographystyle{iclr2023_conference}

\appendix
\section{Appendix: Related Work}
\label{app:related-work}
There are many recent examples of statistical and ML research around digital interventions and mobile health; good examples are \citet{Forman19, Yom17, Liao20}. A data science framework for the design of RL algorithms for digital interventions is presented in \citet{Trella22}. \citet{Wang2021} uses a data-driven behavioral simulator (trained using real data) to model the user's behavior and generate simulated data that can be used to train and evaluate the RL algorithm specifically for mobile health.

The van der Schaar lab has produced seminal and ground-breaking research around synthetic data generated within the framework of privacy~\citep{Jordon2018a, Jordon2019, Jordon2021, Yoon2020}, electronic health record (EHRs), and its application for ML for healthcare~\citep{Alaa2019, Jordon2019, Jordon2020, Jordon2021, Yoon2020}. Their work includes methods to generate data using Generative Adversarial Networks (GANs)~\citep{Jordon2018a, Yoon2019, Yoon2020}, taking into account its time series nature~\citep{Yoon2019, Alaa2021a, Jarrett2021}, evaluation methods for synthetic data generation~\citep{Jordon2018, Alaa2021}, fairness~\cite{Breugel2021}, and methods that include models of some of the physical processes underlying clinical parameters in the generation process~\citep{Alaa2019}.

The KDD Cup Challenge 2019 included a competition on policy learning for malaria control using RL~\citep{Nguyen2019, Zou2021}, for which a global health environment for simulation of RL applications was introduced~\citep{Remy2020}. Understanding the adaptive intervention delivery as a recommendation problem, a simulation environment specifically designed for RL algorithm development and testing is presented in \citet{Ie2019b}. The MARS-Gym framework~\citep{Mars20} provides a set of tools and environments, built on top of the OpenAI Gym library~\citep{Brockman2016}, to model, train and evaluate RL-based recommender systems. In \citet{Singh2020}, a simulation environment is used to test RL recommendation algorithms in the context of safe RL.

\section{Appendix: Adaptive Intervention Delivery Framework}
\label{app:ml-adaptive-intervention}

The architecture for the adaptive intervention delivery framework for mobile health is as follows. The central piece is a data-centric platform that connects to the apps through the \emph{HealthKit} SDK, which ensures all the information that is needed is tracked in a data schema optimized for data science and machine learning purposes. A use-case-specific data pipeline then turns the incoming information into actionable insights and metrics that are then readily available for visualization, and analysis, and to be consumed by (predictive and recommendation) ML models that produce additional metrics. The complete individual characterization provided by the resulting observational, predictive, and recommendation metrics is then available for the design of adaptive interventions. They are used to define the cohort of users that will be targeted, the content and scheduling of the interventions, and the state (or context, as we are limiting the discussion here to the use of contextual and restless bandits) and rewards for the RL component. The latter then uses the information to decide whether to assign subjects to treatment or control groups at each decision point in an adaptive experiment. In the case of adaptive interventions deployed in production, it decides whether to deliver or not or which of a finite set of interventions to deliver at each decision point for all users in the targeted cohort. This is schematically represented in Figure \ref{fig:overall_process} by the orange shaded block labeled \emph{platform} and the cohort of real users it is linked to.

The platform can integrate with the digital tools (and additional sources of digitalized information) through the \emph{HealthKit} SDK, which supports data collection in an opinionated data schema optimized for data science and ML, and which is also responsible for the delivery of content and messaging interventions back to the tool. It guarantees all incoming information into the platform is adequately tracked and organized. 

In the platform, incoming contextual, patient, and FHW data will be processed by use-case-specific data pipelines into useful metrics. All these metrics are then available to run statistical and ML models that will generate additional metrics. Cohorts are defined in terms of these metrics (e.g. FHWs that are predicted to underperform in their next evaluation) and constitute the basic unit of aggregate exploration and intervention. Interventions are defined and can use the available metrics to determine the content and scheduling of the interventions (e.g. content recommendation in the capacity building module as suggested by an ML model, sent in a push notification the next time they are predicted to be online). Intervention delivery is RL-based, and it is these metrics too that can be used to build the reward (e.g. number of questions answered correctly in in-app tests) and define the state for RL algorithms. 

Interventions can combine clinical and behavioral elements. Clinical interventions fall into the realm of precision medicine and involve leveraging all the information available about a patient to improve quality of care and patient prognosis. Behavioral interventions aim at ensuring an adequate level of engagement (with the tool, program, and treatment) of medical care teams and patients. 

Once the interventions have been developed, their effect can be measured using the experimentation platform, which includes fully randomized and adaptive (bandit-based) designs, with and without multiple assignments.


\section{Appendix: Mathematical Details}
\label{appendix:maths}
User behavior is defined using a decay model on the activity  concerning the number of nudges $n\geq0$ received on the last $d$ days ( $d=5$ in our experiments). The decay model is represented by the three equations, each describing different aspects of the user behavior:

        \begin{equation}
            \label{eq:decay_1}
            f(n) =  a_0 e ^{-k_an}
        \end{equation}
        \begin{equation}
            \label{eq:decay_2}
            g(n) = b_0\frac{k_a\left(e^{-k_an} - e^{-k_bn} \right)}{k_b-k_a}
        \end{equation}
        \begin{equation}
            \label{eq:decay_3}
            h(n) = c_0\frac{k_a\left(e^{-k_bn} -1\right) - k_b\left(e^{-k_an} -1\right)}{k_b-k_a}
        \end{equation}
        where $k_a$, $k_b$, $a_0$, $b_0$, $c_0$ are parameters defining the shape of the function. These parameters can be determined through a fitting process using real-world historical data to ensure that the decay model accurately reflects users' actual behavior. The decay model is then used to generate synthetic user behavior data for testing reinforcement learning algorithms in the context of e-health interventions.
        
        Each  equation models a different type of behavior: eq. (\ref{eq:decay_1}) models the users whose maximum interaction with the app is when they do not receive intervention, eq. (\ref{eq:decay_2})  represents users that are reactive to the interventions but are eventually overwhelmed and stop interacting. Finally, eq. (\ref{eq:decay_3}) models users that are reactive to interventions, but at some point, they are saturated, and the activity does  not increase  or decrease. Figure \ref{fig:modelBehaviour} shows the shape of the three decay models.

        The final behaviour $a_i(n)$ for user $i$ is a linear combination of $f,g,$ and $h$:
        \begin{equation}
        \label{eq:final_action}
            a(n) = \alpha_i f(n) + \beta_i g(n) + \gamma_i h(n)  
        \end{equation}
        where $\alpha_i$, $\beta_i$, $\gamma_i$ are parameters defining the individual characteristic of user $i$ and can be itself random variables.

        This allows us to define behavior at three levels, to which we will refer as  \emph{individual}, \emph{contextual}, and \emph{intervention}. By \emph{individual}, we mean the unique behavior of a user as an individual, independent of the features defining their context. \emph{Contextual} behavior represents the expected behavior of a user based on their context. All users with the same context will exhibit similar behavior to some extent, which can also be regulated. This is achieved using different parameters or even different functions $f(n),g(n),h(n)$ depending on the context. Finally, the user changes in in-app activity in response to an \emph{intervention} depend on the number of nudges $n$ received over time. It is implemented using the decay model described above, where the number of nudges $n$ directly depends on the assigned group of the previous decision points, i.e., eq. (\ref{eq:decay_1})-(\ref{eq:decay_3}) naturally models this behaviour. Note that it is also possible to define user-specific reactions to the interventions in the app's level or type of activity. This is achieved using  different weighting parameters  $\alpha_i$, $\beta_i$, $\gamma_i$ for each user $i$.

\end{document}